# A wearable anti-gravity supplement to therapy does not improve arm function in chronic stroke: a randomized trial

## AUTHORS

Courtney Celian, M/OT[a], Partha Ryali, PhD[a,b], Valentino Wilson, MEng[a,b], Adith Srivatsa[a,b], James L. Patton, PhD[a,b]

[a] Shirley Ryan AbilityLab, Chicago, IL

[b] Department of Biomedical Engineering, University of Illinois at Chicago, Chicago, IL

## CORRESPONDING AUTHOR

James Patton, PhD, Professor, University of Illinois at Chicago, Senior Research Scientist, Shirley Ryan AbilityLab, 355 E. Superior St, Chicago, IL 60611   pattonj@uic.edu   +1 312 413 7664

## ABSTRACT

**Background:** Gravity confounds arm movement ability in post-stroke hemiparesis. Reducing its influence allows effective practice leading to recovery. Yet, there is a scarcity of wearable devices suitable for personalized use across diverse therapeutic activities in the clinic.

**Objective:** In this study, we investigated the safety, feasibility, and efficacy of anti-gravity therapy using the ExoNET device in post-stroke participants.

**Methods:** Twenty chronic stroke survivors underwent six, 45-minute occupational therapy sessions while wearing the ExoNET, randomized into either the treatment (ExoNET tuned to gravity-support) or control group (ExoNET tuned to slack condition). Clinical outcomes were evaluated by a blinded-rater at baseline, post, and six-week follow-up sessions. Kinetic, kinematic, and patient experience outcomes were also assessed.

**Results:** Mixed-effect models showed a significant improvement in Box and Blocks scores in the post-intervention session for the treatment group (effect size: 2.1, p = .04). No significant effects were found between the treatment and control groups for ARAT scores and other clinical metrics. Direct kinetic effects revealed a significant reduction in muscle activity during free exploration with an effect size of (-7.12%, p< 005). There were no significant longitudinal kinetic or kinematic trends. Subject feedback suggested a generally positive perception of the anti-gravity therapy.

**Conclusions:** Anti-gravity therapy with the ExoNET is a safe and feasible treatment for post-stroke rehabilitation. The device provided anti-gravity forces, did not encumber range of motion, and clinical metrics of anti-gravity therapy demonstrated improvements in gross manual dexterity. Further research is required to explore potential benefits in broader clinical metrics.

This study was registered at ClinicalTrials.gov (ID# NCT05180812).

# INTRODUCTION

Each year, approximately 800,000 individuals experience a stroke, leading to significant disability in upper extremity movement and accumulating healthcare expenses that exceed $3 billion. Anti-gravity therapy through mobile arm support is commonly used by occupational therapists to aid stroke survivors in regaining upper extremity movement and function. These devices counteract the effects of hemiparesis, or weakness on one side of the body, by partially or fully unweighting the affected arm. This reduced loading allows patients with diminished strength to actively participate in recommended stroke rehabilitation activities like range of motion exercises and strengthening.[1] Moreover, these devices enable greater freedom of movement at the shoulder joint while discouraging compensatory motions such as excessive trunk movements or abnormal positioning of the shoulder girdle. These compensatory movements can lead to muscle imbalances, impede arm recovery, and increase risk of further injury if not addressed properly.[2–4]

Anti-gravity therapy increases accessible range of motion and mitigates the effects of abnormal muscle synergies that commonly develop post-stroke.[5–7] This facilitates more natural movement patterns and creates a better environment for motor re-learning and recovery. However, many available commercial arm support devices are expensive, ranging from $300 to over $6,000. They are bulky and limit use to table-top activities since they require mounting to an immovable surface or frame. This significantly reduces their versatility and flexibility for applying arm support during functional, occupation-based treatments.

Robotic solutions have proven to be safe alternatives, eliciting intensive movement training, improved coordination, and facilitating active movement.[8–13] However, effect sizes tend to be small, show little clinical significance, and lack functional relevance and flexibility in training due to the pre-defined paths that may not align with functional contexts.[9] To date, there are no solid results to show which robotic system is the most effective since outcome depends on the patient's clinical state. However, it is advisable to use systems favoring the subject's active participation in movement if the subject's motor conditions permit.[14] In fact, functional recovery is achieved through use-dependent cortical reorganization and active participation favors increased levels of physiological reorganization and provides more clinical benefits.[15–17]

Here we present the ExoNET —a passive robotic technology designed to counteract gravitational forces on the upper extremities. Employing personalized spring networks, the ExoNET not only simplifies the training regimen but may offer clinically significant benefits for those with neurological impairments. In contrast to current commercialized anti-gravity devices, the ExoNET is directly attached to the user, providing enhanced support in any functional position. The introduction of the ExoNET aims to bridge the gap between the need for effective rehabilitation and the practical constraints of existing solutions. By focusing on anti-gravity therapy, this study explores a novel avenue to enhance stroke recovery through a wearable device.

Through the ExoNET system, we aimed to assess the safety, feasibility, and efficacy of this new form of anti-gravity therapy tailored for a post-stroke population. Specifically, we investigated whether chronic stroke survivors undergoing occupational therapy (OT) focused on the remediation of upper extremity movement with an anti-gravity ExoNET, would improve across clinical outcome measures. We hypothesized that conventional OT-facilitated upper extremity treatment combined with anti-gravity support of the ExoNET would lead to higher gains in the Action Research Arm Test (ARAT).[18] To verify the safety and feasibility of the device, we measured bicep muscle activity and movement kinematics.

## METHODS

This study was a double-blind, randomized controlled clinical trial. The Northwestern University Institutional Review Board (Chicago, IL) approved this study (STU00216062) and served as the IRB

Page **4** of **28**

of record; reliance was established with the University of Illinois at Chicago research ethics authority (STUDY2021-1492). Prior to commencing the study, all subjects provided written informed consent in adherence to the Declaration of Helsinki.

This study was completed at the Shirley Ryan AbilityLab (SRALab) rehabilitation hospital; a translational hospital, where clinicians, patients, scientists, and engineers meet and do research within the clinical treatment areas. Clinical evaluations took place in SRALab Robotics Lab area in the Center for Neuroplasticity. All treatments took place on the floor adjacent to the Legs and Walking AbilityLab, which was done to mimic a real-world clinical setting for our post-stroke population.

Study subjects were recruited by the study team members from SRALab's stroke research registry and from flyers posted around SRALab. All eligible subjects were older than 18, at least 8 months post-stroke, hemiparetic but with some degree of both shoulder/elbow movement capabilities, with a moderate impairment level determined by a Fugl-Meyer Upper Extremity (FMUE) score between 15-50. Individuals were excluded from the study if they had bilateral paresis; shoulder pain/articular rigidity; spasticity of the upper extremity in rigid flexion or extension; aphasia, cognitive impairment, or affective dysfunction that would influence the ability to perform the experiment; or recent upper extremity Botox injections (within 4 months). This study excluded individuals under the age of 18, pregnant women, prisoners, and cognitively impaired adults.

### Device

The ExoNET is a passive wearable device designed and built at the Robotics Lab of the SRALab (fig. 1). It is composed of adjustable, multi-joint elastic elements which, combined in parallel, can provide any desired sagittal-plane torque to the upper extremity. The idea is based on the extension of the MARIONET concept, a diagonal tension element able to exert a desired torque at any joint[19], into a system of parallel MARIONETs to obtain any desired torque, by linearly summing the torques generated by each element. As a result, the ExoNET can generate highly customizable, multi-joint, upper extremity torque profiles and is also able to provide gravity compensation, whose goal is to reduce the weight of the arm joints during motion.[20] Along with its customizability, the ExoNET is versatile in therapeutic applications across a range of contexts and positions, enabling users to harness anti-gravity therapy for diverse goals.

---

***INSERT FIGURE 1 ABOUT HERE** (device)*

---

## Clinical Outcome Measures

Clinical outcome measures were assessed before the intervention at week 1 (baseline), after the intervention at week 4 (post), and approximately 6 weeks after the intervention (follow-up). The primary outcome measure was the Action Research Arm Test (ARAT)[18], a 19-item measure that assesses upper extremity function across 4 domains (grasp, grip, pinch, and gross arm movement). Performance on each item is scored on a 4-point ordinal scale ranging from 0 to 3. Scores are summed, with a possible range of 0 to 57, and a higher score indicating higher levels of function in the upper extremity.

Secondary clinical outcomes included the Box and Blocks Test (BBT)[21], an assessment of unilateral gross manual dexterity consisting of 1.50-2cm blocks and a 53.7cm x 25.4cm wooden box divided by a 15.2cm high partition creating two compartments. The score was determined by how many blocks the subjects moved from one side of the box to the other in 60 seconds.

The Fugl-Meyer Upper Extremity assessment (FMUE)[22] evaluated change in upper extremity hemiplegic motor impairment. The FMUE evaluates gross and fine motor control, 5 common grasp positions, and reflex activity of the elbow and long finger flexors. It is comprised of 33 items, using a 3-point ordinal scale from 0 to 3 where a lower score denotes greater impairment. The maximum score is 66.

The Wolf Motor Function Test (WMFT)[23] is a 17-item assessed upper extremity motor ability through timed and functional tasks. Each item is timed, with 120 seconds being the maximum time allotted. Each functional task uses a 6-point ordinal scale from 0 to 5 where a lower score indicates a lower functioning level. The maximum score is 75.

### Muscle Activity and Movement Kinematics

To measure the impact of the ExoNET's gravity support, we calculated kinetic outcomes through direct and longitudinal assessments of muscle activity at the long heads of the biceps using Electromyography sensors (EMG) (Delsys, MA, USA). To analyze movement kinematics, we evaluated the direct and longitudinal changes in free exploration position and velocity coverage, calculated using an Xbox Kinect (Microsoft, WA, USA) joint tracking system on wrist position.

### Subject Experience

We use the Intrinsic Motivation Inventory (IMI)[24] to capture subjects' subjective experience using the ExoNET. The IMI was developed to assess across 7 dimensions; however, we adapted the tool to measure across 5 different dimensions that applied to our device: effort/importance, value/usefulness, interest/enjoyment, perceived competence, and pressure/tension.

### Study Procedures

This study compared the outcomes between subjects who received conventional occupational therapy (OT) intervention while wearing the ExoNET upper extremity device tuned to either anti-gravity support (treatment group), or no gravity support (control group). All subjects engaged in six, 45-minute OT treatments 2-3 times a week across 2-3 weeks, and completed baseline, post, and follow-up evaluations. See supplementary figure 1 for an outline of the study protocol.

Subjects were consented, and eligibility was determined via an in-person screen by the study project coordinator who is also a licensed and registered occupational therapist. Eligible subjects were randomized into either the treatment (ExoNET tuned to gravity-support) or control group (ExoNET tuned to slack condition) using a 1:1 allocation ratio. Subject group randomization was conducted using a random number generator in Python, where each subject was randomly assigned to either the number 1 (treatment group) or number 2 (control group) until there were 10 subjects in each group. Randomization and allocation of subjects to either the treatment or control groups was carried out by a lead researcher involved in the study. All subjects completed baseline clinical assessments (ARAT, FMUE, WMFT, and BBT) with a blinded, licensed OT. Kinetic and kinematic outcomes were collected during a 2-minute, free exploration task performed between the ARAT and WMFT. Subjects' goals were discussed and recorded using the Canadian Occupational Performance Measure (COPM), a semi-structured interview tool designed to pinpoint the subject's self-assess occupational performance across self-care, productivity, and leisure domains.[25]

Each session lasted 90-minutes and comprised of 2 phases: kinetic/kinematic data collection, and occupational therapy. During the first part of the session, kinetic and kinematic data were systematically collected in 2-minute intervals under three distinct conditions: (1) without the ExoNET, (2) with the ExoNET donned with no springs, and (3) device donned with springs calibrated to gravity support or slack, based on the subject's group assignment. In the second half of the session, subjects participated in 45-minutes of goal-oriented OT while wearing the ExoNET. The session concluded with another 2-minute free exploration task with the device donned and the springs calibrated based on the subject's group assignment.

Outcome measures, including clinical assessments, kinetic and kinematic data, and patient-reported experiences, were evaluated one week after the final intervention session (post evaluation) and again 5-6 weeks later (follow-up evaluations).

## Statistical Analysis

We used mixed-effects models to examine the efficacy of the anti-gravity therapy on clinical outcomes. For all clinical outcome measures, the model considered the dependent variable as the change in clinical score from baseline to either post or follow up, and the direct and interaction effects of intervention type and session. The 'subject' served as the random effects variable to account for inter-subject variability.

We used mixed-effects modeling to investigate direct and longitudinal kinetic and kinematic effects. For direct effects, the models assessed condition-specific influences while accounting for random effects attributed to individual subjects during their initial treatment session. We analyzed longitudinal effects by considering the interaction between group, session, and time during the no device free exploration task across all treatment days, again treating subjects as a random effect.

# RESULTS

## Recruiting and Enrollment

This study was conducted from March 2021 to May 2022. Thirty-seven (n=37) subjects with chronic symptoms following stroke were considered for enrollment (fig. 2). Each were evaluated for eligibility and 23 subjects met the required criteria. Two subjects withdrew during the study due to scheduling conflicts, and 1 subject was removed by researchers due to unanticipated device modification, rendering data incomparable. Twenty (n=20) subjects (52 ± 13 years) completed all interventions and post-evaluations, but one was lost to follow-up. Table 1 in the supplementary section details all the demographic characteristics and the group allocation for all 20 subjects (including the 1 subject lost to follow-up). The study ended when the enrollment goal of 20 participants was reached.

---

**INSERT FIGURE 2 ABOUT HERE** *(Consort)*

---

## Clinical Outcomes

The result of our primary hypothesis was that ARAT was unaffected by anti-gravity therapy (treatment group-by-time interaction). Individual subject trends were variable (fig. 3). Although the treatment group improved from baseline to post, this was not significant (effect size = 1.3, p = .075) (fig. 3). Paradoxically, our secondary metric, BBT, increased an average of 2.1 from baseline to post for only the treatment group (p = .04), but this was not superior to the control group (Tukey Post-hoc p > .05). WMFT and FMUE did not demonstrate notable changes. Hence while there was evidence of improvement due to therapy, there was no evidence of superiority due to anti-gravity.

---

**INSERT FIGURE 3 ABOUT HERE** *(clinical scores)*

---

## Neuromechanical Effects

Despite the absence of evidence supporting anti-gravity, the device did influence muscle activity. When individuals donned the device but did not engage the springs (slack condition), there was an increase in muscle activity (fig. 4 A and C). When engaging the springs, biceps muscle activity reduced by an average of 7.12% (p<.005) for subjects in the treatment group when they performed free exploration tasks with anti-gravity support compared to free exploration with no device. The reduction in muscle activity indicated an effective reduced effort from anti-gravity assistance. There were no longitudinal trends in this reduction across training days.

---

*\*\*INSERT FIGURE 4 ABOUT HERE\*\* (EMG)*

---

In contrast, we did not observe kinematic changes in mobility; either immediately when donning the devices (within the first day of treatment), or longitudinally (across days) as a consequence of our intervention (fig. 5). This was measured by joint range coverage, and similar to previous tests on individuals without neural injury [26], where individuals distributed their motions similarly with or without the device. Our failure to observe any changes in mobility suggests that the device did not encumber the participants, and therapy motions could continue without a loss of motion range.

---

*\*\*INSERT FIGURE 5 ABOUT HERE\*\* (Kinematic measures)*

---

## Subject Experience

The Intrinsic Motivation Inventory (IMI) results indicate a high level of engagement with the ExoNET (fig. 6), with participants in the treatment group showing a slightly higher effort/importance score (4.70) compared to the control group (4.56). While the treatment group's interest/enjoyment score (4.69) was slightly lower than that of the control group (4.88), it indicates a broadly positive experience with the device. Competence levels reported were similarly high in both groups (control: 4.97, treatment: 4.98), suggesting that users across groups felt adept in using the ExoNET. Notably, the treatment group reported lower pressure/tension (2.52) compared to the control group (3.10), which may point to a less stressful experience with the ExoNET. Value/usefulness was rated highly by both groups, with the treatment group (6.35) rating it slightly higher than the control group (6.25). These observations highlight that despite the lack of significant changes in clinical scores, the ExoNET is perceived as a positive element in therapy, with potential benefits as identified by users themselves.

***INSERT FIGURE 6 HERE** (IMI)*

## DISCUSSION

Anti-gravity as an adjunct to upper extremity therapy was found to be safe and feasible. This pilot was limited to individuals with chronic movement deficits post-stroke and employed an ExoNET passive robot that was easy to use, lightweight, and wearable. Muscle activity associated with wearing the device unengaged was higher, but lower with anti-gravity engaged. The anti-gravity group did not demonstrate superior improvements in the clinical outcome measures.

There was evidence of some improvements, however. One measure, BBT, improved significantly for the anti-gravity group from baseline to post evaluations, but the control group did not. Anti-gravity improvements in gross manual dexterity and object transfer abilities during task-specific activities. However, the improvement demonstrated by the treatment group was not significantly better compared to the control group. Additionally, the lack of significance on the ARAT indicates that more comprehensive upper extremity function may require a longer treatment duration or higher dosage.

Failing to show a clinical superiority for anti-gravity did not mean that gravity assistance was not delivered. We found significantly reduced bicep muscle activity in the treatment group when the device was engaged. Reducing excessive muscle strain could promote longer exercise tolerance, reduce compensatory movements, make the patient more motivated, and increase available range of motion for carryover to real-world tasks. While this study examined an anti-gravity application, the spring configuration can theoretically provide diverse therapeutic forces beyond gravity support. These include resistance for strength training and error-augmentation for motor learning.[27,28] Future work should explore these alternative modes to further elucidate the ExoNET's rehabilitation potential.

Importantly, there was no evidence that anti-gravity treatment limited participants' range of motion. We found no significant changes in position and velocity coverage in either the treatment or control group with the device. Subjects performed a variety of functional activities across multiple positions (e.g., seated, standing, supine, and side-lying) without hindering their existing capabilities. This is usually the main drawback of current iterations of commercially available anti-gravity devices, which must be mounted on a cart, table, or wheelchair. In contrast, the ExoNET is wearable, generalizable to a variety of tasks and positions. Such real-world use during functional tasks warrants examination of its efficacy as an assistive technology for enhancing skill and consequently independence.  Enhancing activity can bring new skills, and new skills bring more activity.

Subjective feedback indicated that therapy was well-tolerated, with subjects reporting high interest/enjoyment and low perceived tension on the IMI. Most importantly, all subjects perceived value and usefulness, despite our not detecting significant improvement in our clinical scores. However, more systematic feedback and survey techniques are needed to determine both participants' and therapists' perceptions and decision making. Here, a singe therapist was used throughout for consistency, but other studies might test reliability with several. Interestingly, all subjects reported that they contributed effort but not energy into therapy. It may be that therapy involves a mental contribution related to effort.

Overall, anti-gravity training using a wearable ExoNET shows can be an adjunctive to upper extremity stroke rehabilitation. Such devices are highly versatile in the mechanical "program" that can be put in. allow unrestricted movement compared to most static systems and are customizable for therapeutic forces beyond anti-gravity, including resistance, bistable attractors, and error augmentation.[28] Further investigation into optimal treatment parameters, longer term outcomes, and diverse clinical applications is merited.

This study was not without its limitations. Sample size was modest, and a follow-up period of six weeks might not be sufficient to capture long-term effects. The lack of significant results in ARAT scores suggests that other outcome measures may be necessary to adequately capture functional improvement. We only assessed the subject's upper extremity functional capacity using the ARAT, not upper extremity performance in daily life, so it remains to be seen if anti-gravity support may provide benefit gauged in some other manner measured using different metrics.[29,30] COPM was used to gauge subjects' goals, but no follow up was completed, missing important clinical metrics of subjects perspectives of outcome assessment. Good IMI scores might be driven by social and psychological effects and the will to please. The device was donned by the researcher, and in its current design iteration, cannot be donned by the user alone. Results from this study were consistent with literature on robot-assisted therapy, which found that patient-specific factors considerably influence the outcomes of such interventions.[31] Using this in an acute setting may find higher benefit, since improvement in chronic stroke population are incremental and occur over extended periods of time.

## CONCLUSIONS

This double-blind, randomized controlled clinical trial demonstrated the safety, feasibility, and clinical effects of anti-gravity therapy during OT treatment in chronic stroke. This study showed some improvements in clinical metrics, but the anti-gravity supplement did not prove to be superior. Measurements of kinetics and kinematics revealed that gravity support reduced effort without reducing arm movement range. These results were obtained during standard rehabilitation practice in a real-world therapy environment. Such convenient wearable systems may be better suited for simple assistance of function rather than therapy. However, as therapy often requires some movement ability, such assistance may benefit individuals Such low function that they could not otherwise move. Other settings for such devices, which can deliver a wide array of forces, may be more beneficial.

## ACKNOWLEDGEMENTS

We thank the members of the RobotLab at the Center for Neuroplasticity at the Shirley Ryan AbilityLab for their weekly insights and comments. We would especially like to acknowledge Yaseen Ghani, Beatrice Malizia, and Jialin He for their support in data collection. We thank the clinicians in the Arms & Hands lab at the Shirley Ryan AbilityLab for their encouragement and their constructive criticisms at the early stages of design in our Innovation Huddle. We also thank our collaborators from the COMET RERC and Shirley Ryan AbilityLab administration for their insights and feedback to improve the ExoNET for clinical adoption. This work was supported by the National Institute on Disability, Independent Living, and Rehabilitation Research [90REGE005] Collaborative Machines Enhancing Therapy (COMET) RERC.

## DATA AVAILABILITY STATEMENT

The data collected and analyzed throughout this study are accessible via the following Dropbox link:

https://www.dropbox.com/scl/fo/vez0pkhpfjiubxgy7wib4/AKXDX6eRvRQ53QD_xjRnmB0?rlkey=r4x95028jrwx65u47f26zou9m&dl=0

## DECLARATION OF CONFLICTING INTERESTS

The authors have disclosed that they have no conflicts of interest regarding the research, authorship, and publication of this article.

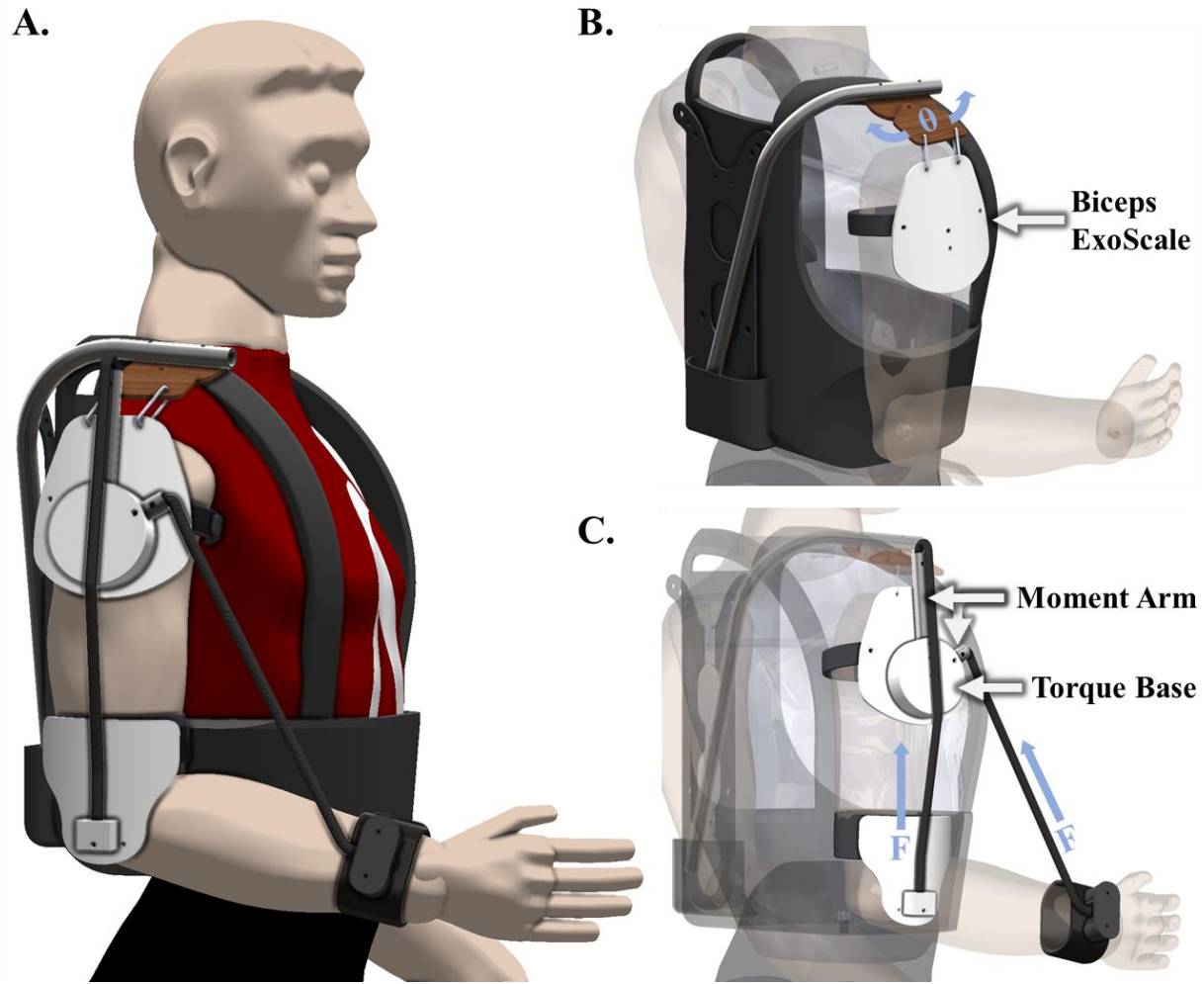

**Figure 1. ExoNET Upper Extremity Device.** The device (A) is composed of two sub-assemblies: (B) shoulder-back-biceps attachment and (C) torque delivering system. The shoulder-back-biceps attachment sub-assembly provides rigid connection to the body while maintaining full range of motion at the shoulder joint, constraining the gravity compensation torque field to sagittal plane movements. The torque delivering system attaches to the shoulder through torque bases and moment arms, allowing multi-joint elastic element attachment between the biceps-elbow and biceps-wrist.

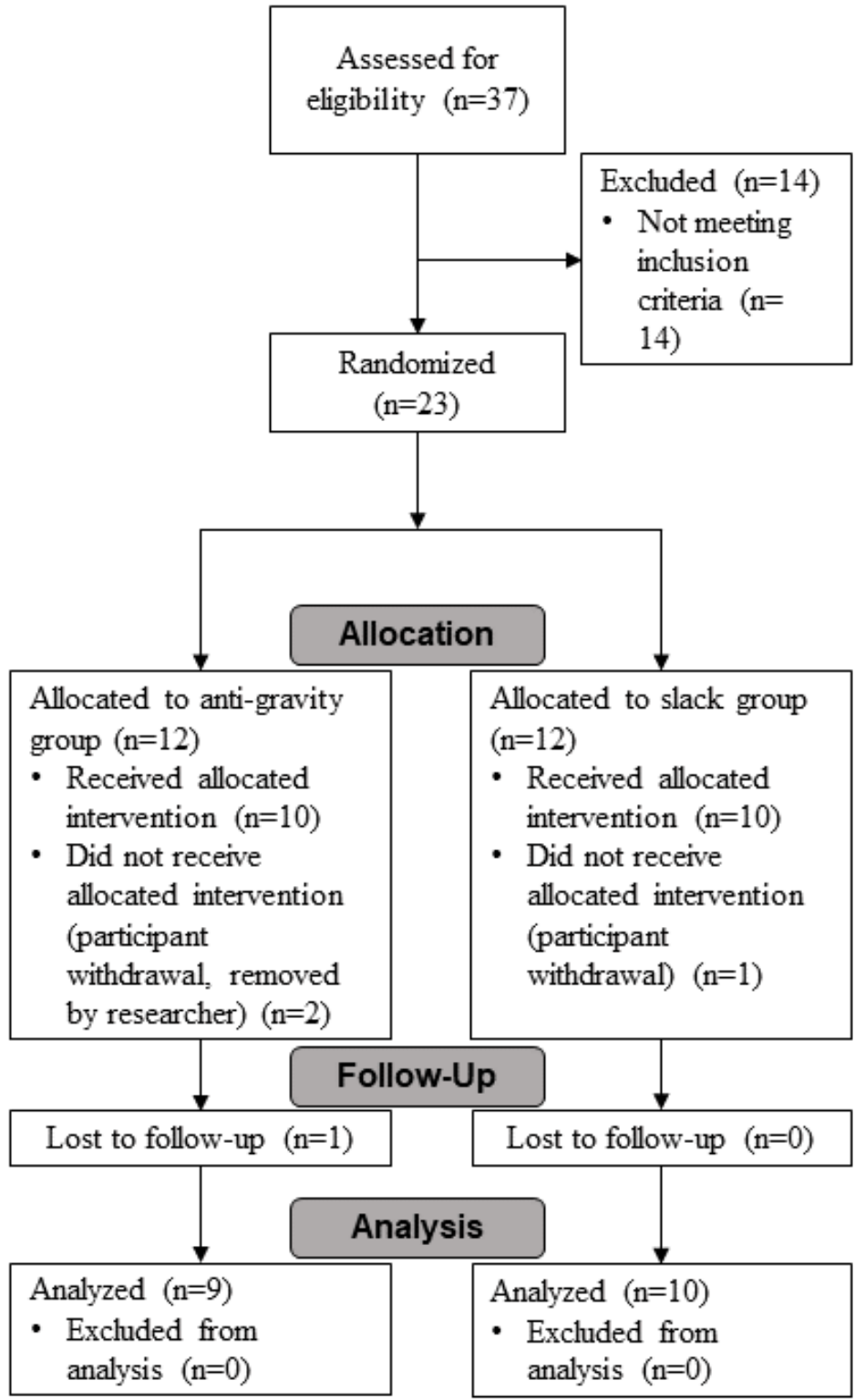

**Figure 2**. Participant recruitment process diagram, following the guidelines outlined in the CONSORT protocol.[31]

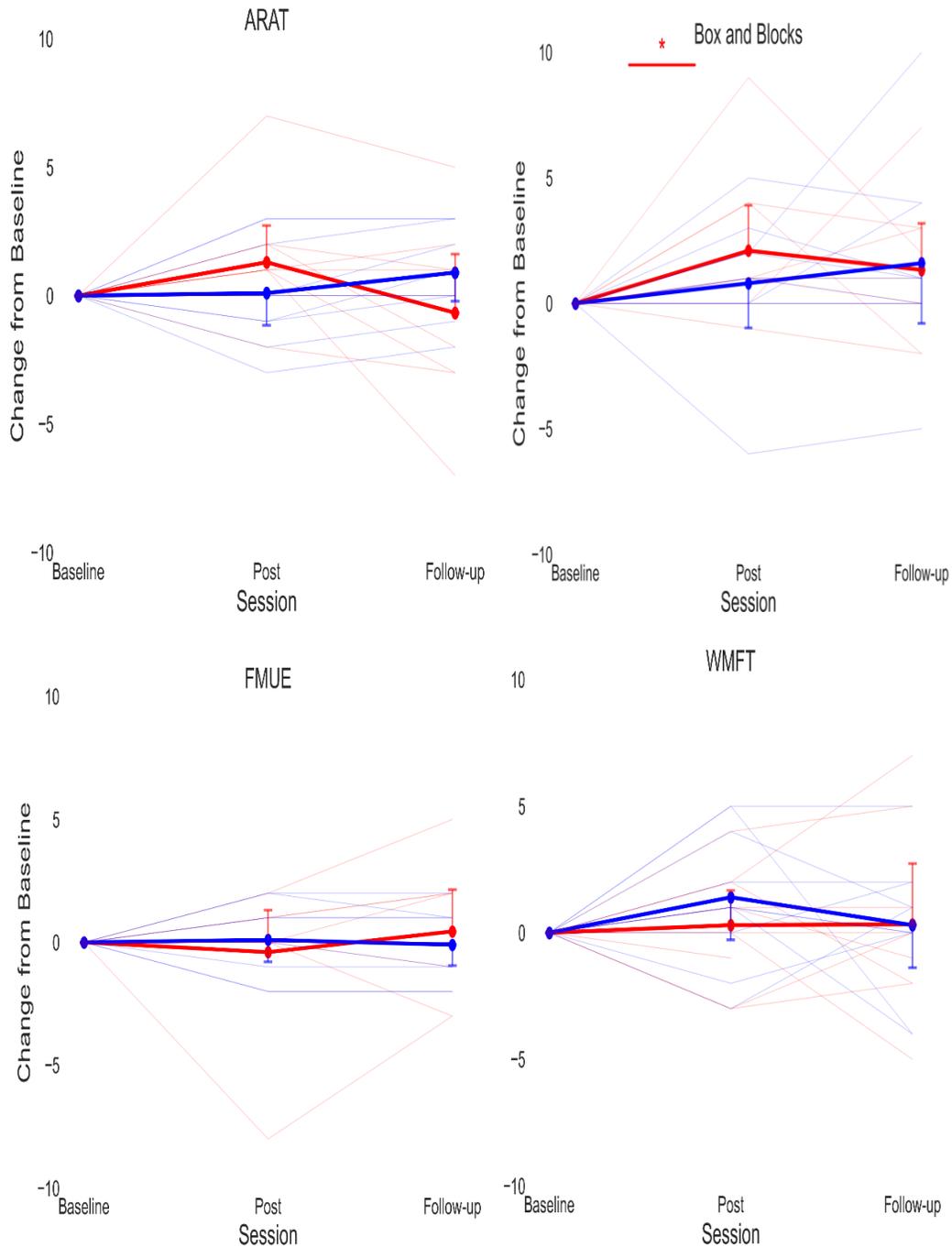

**Figure 3.** Our primary outcome measure, ARAT, is shown with 3 secondary clinical measures. BBT showed significant improvement from baseline to post (p = .04). Mixed effect modelling revealed ARAT had a non-significant increase (session effect) from baseline to post with anti-gravity intervention (effect size = 1.3, p = .075). BBT had a significant session effect from baseline to post with the anti-gravity intervention (effect size = 2.1, p < .05). All other group, session, and interaction effects were negligible and non-significant.

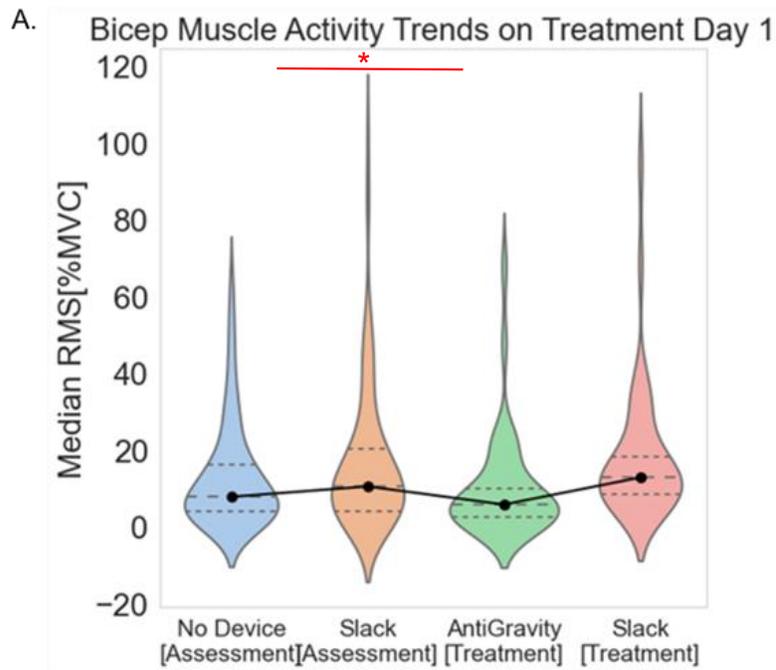
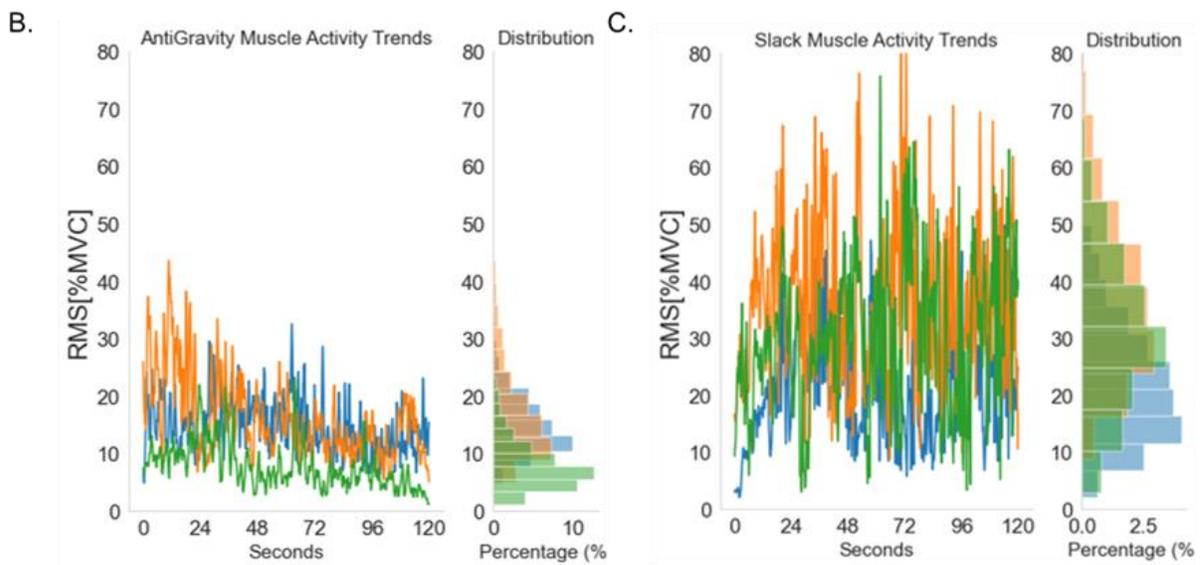

**Figure 4. Direct Kinetic Effects of Anti-Gravity and Sham Control ExoNET on Post-Stroke Muscle Activity** (A) Bicep activity trends for 2 minutes of free exploration in various device conditions; median in black. Anti-gravity had a -7.12% rms[%mvc] decrease in biceps ($p < .005$), compared to the sham "slack" treatment that experienced no change. Violin widths depict the activity distribution across conditions, highlighting a tendency for a higher and broader range of activity for the slack group. (B & C) Example direct effects on activity for a single subject in each group during free exploration where they were continuously moving. Colors correspond to the conditions shown in (A). All plots show a reduction in activity when using the anti-gravity device, and a heightened activity with the sham slack condition.

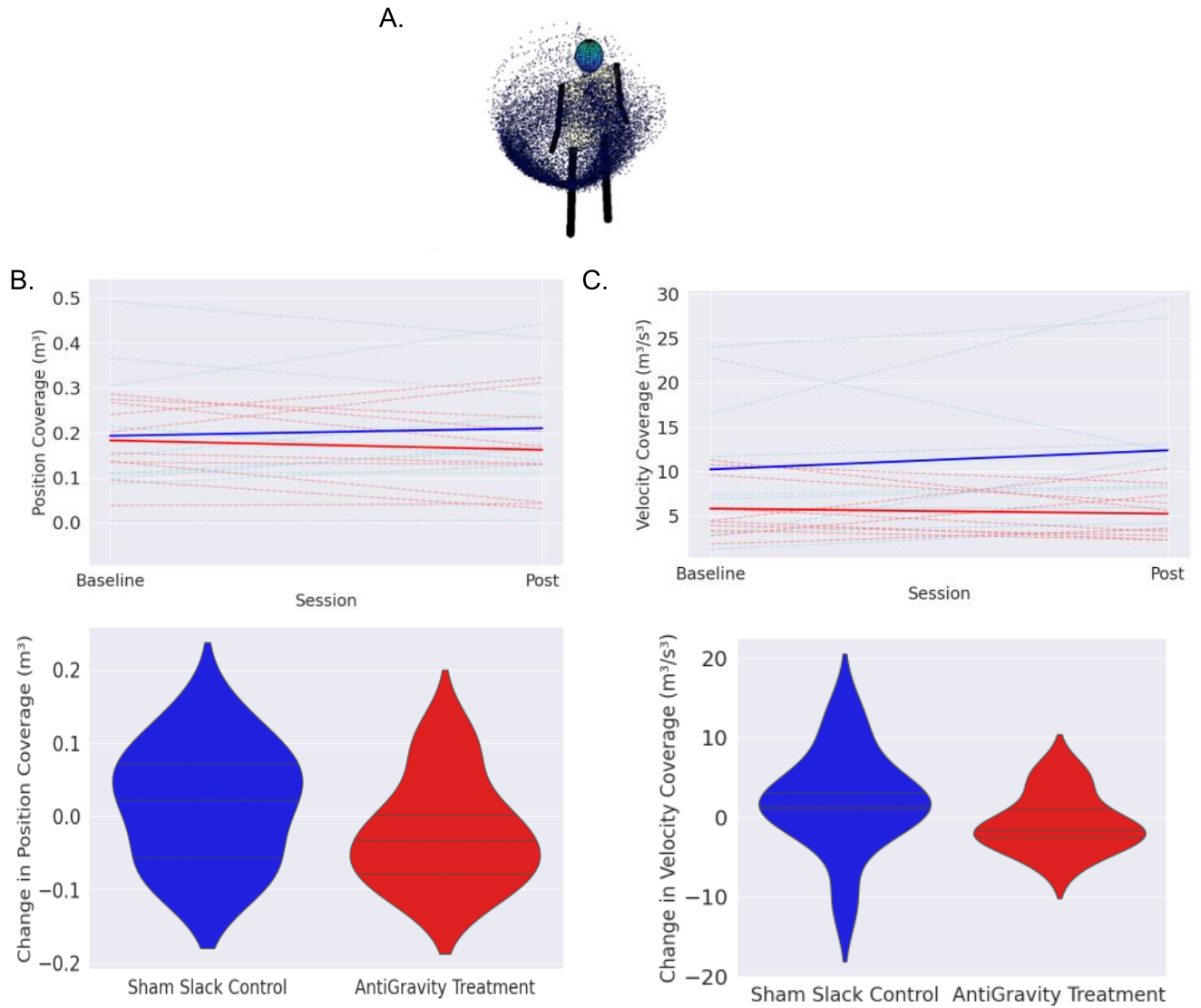

**Figure 5 – Kinematic Effects in Position and Velocity Coverage.** (A) Example Distribution of wrist positions during 2 minutes of free exploration. (B) Position coverage before and after intervention for sham slack Control (blue) and anti-gravity Treatment (red), with individual and mean trends, alongside violin plots depicting distribution of changes. (C) Velocity coverage trends and distribution changes. There were no significant kinematic differences observed across time or device setting.

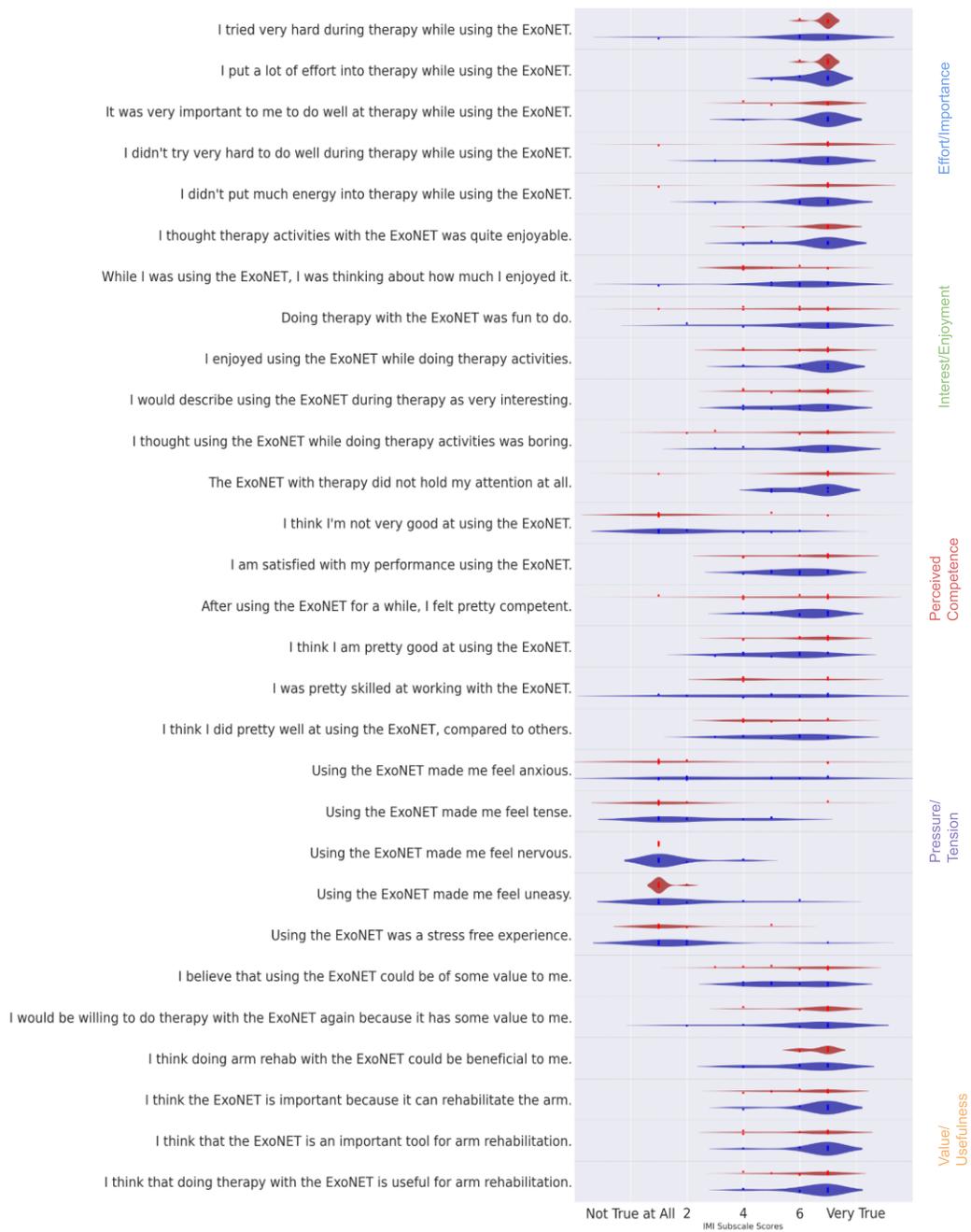

**Figure 6** – Intrinsic Motivation Inventory (IMI) scores for ExoNET in post-stroke.

# APPENDIX

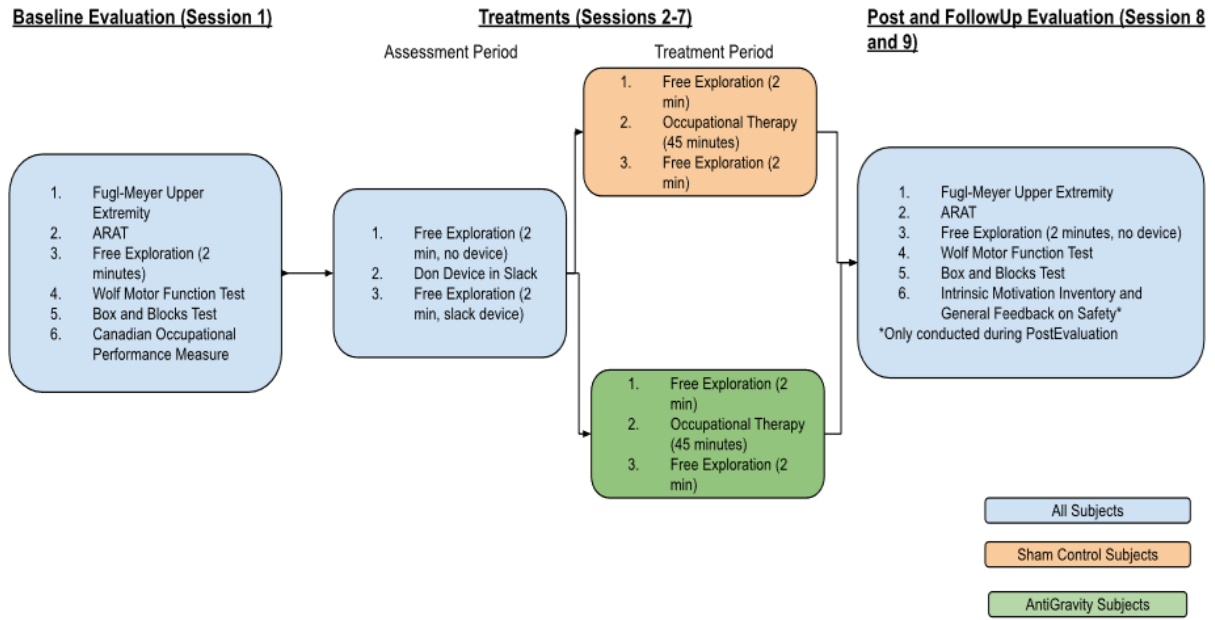

**Supplemental Figure 1**: Outline of the study protocol.

| Characteristics | Treatment Group (n=10) | Control Group (n=10) | Total (n=20) |
| --- | --- | --- | --- |
| **Sex, n(%)** | | | |
| Male | 6 (60%) | 7 (70%) | 13 (65%) |
| Female | 4 (40%) | 3 (30%) | 7 (35%) |
| **Age (Mean ± SD)** | | | |
| Years | 57 ± 10 | 47 ± 13 | 52 ± 13 |
| **Time from Stroke (Mean ± SD)** | | | |
| Months | 82 ± 41 | 44 ± 27 | 62 ± 39 |
| **Hemiparesis, n(%)** | | | |
| Right | 5 (50%) | 3 (30%) | 8 (40%) |
| Left | 5 (50%) | 7 (70%) | 12 (60%) |
| **Handedness, n(%)** | | | |
| Right | 10 (100%) | 9 (90%) | 19 (95%) |
| Left | 0 | 1 (10%) | 1 (5%) |
| **Stroke Type, n(%)** | | | |
| Ischemic | 7 (70%) | 3 (30%) | 10 (50%) |
| Hemorrhagic | 3 (30%) | 4 (40%) | 7 (35%) |
| Unknown | 0 | 3 (30%) | 3 (15%) |

**Supplemental Table 1.** Demographic characteristics of all subjects who received allocated intervention.